\documentclass{article}

\usepackage{PRIMEarxiv}

\usepackage[utf8]{inputenc} 
\usepackage[T1]{fontenc}    
\usepackage{url}            
\usepackage{booktabs}       
\usepackage{amsmath,amssymb,amsfonts}
\usepackage{nicefrac}       
\usepackage{microtype}      
\usepackage{fancyhdr}       
\usepackage{graphicx}       
\usepackage{subcaption}
\usepackage{algorithm}
\usepackage{algpseudocode}
\algnewcommand{\LeftComment}[1]{\textit{\Statex \# #1}}
\usepackage{comment}
\usepackage{wrapfig}
\usepackage{xcolor}
\usepackage[numbers,sort&compress]{natbib}
\definecolor{linkblue}{rgb}{0.21,0.49,0.74}
\usepackage[breaklinks,colorlinks,allcolors=linkblue]{hyperref}

\title{CRLoMTL: Conflict-Resolution through Low-rank matrices for Multi-Task Learning}

\author{
  Ahmed Agiza\textsuperscript{*} \quad
  Tasneem Shaffee\textsuperscript{*,\ensuremath{\ddagger}} \quad
  Rahil Badkul \quad
  Sherief Reda \\
  School of Engineering \\
  Brown University\\
  Providence, RI, USA \\
  {\small\texttt{ahmed\_agiza@brown.edu} \quad
  \texttt{tasneem\_shaffee@brown.edu} \quad
  \texttt{rahil\_badkul@brown.edu} \quad
  \texttt{sherief\_reda@brown.edu}}
}

\begin{document}
\maketitle

\makeatletter
\begingroup
\renewcommand{\thefootnote}{\fnsymbol{footnote}}
\renewcommand{\@makefntext}[1]{%
  \noindent
  \makebox[1.2em][l]{\@makefnmark}#1%
}
\footnotetext[1]{Equal contribution.}
\footnotetext[2]{Code: \href{https://github.com/scale-lab/CRLoMTL.git}
{\texttt{github.com/scale-lab/CRLoMTL}}}
\footnotetext[3]{Tasneem Shaffee is on leave of absence from Ain Shams University, Cairo, Egypt.}
\endgroup
\makeatother

\begin{abstract}

Multi-task learning (MTL) improves model efficiency by sharing representations across related vision tasks, but it often suffers from negative transfer when task updates interfere with one another. Existing conflict-mitigation methods typically operate in gradient space, requiring explicit task-gradient manipulation and additional optimization steps. We propose \textit{Conflict-Resolution through Low-Rank Matrices for Multi-Task Learning (CRLoMTL)}, a parameter-efficient framework that mitigates task interference in the low-rank parameter-update space. CRLoMTL learns task-specific LoRA updates through task-specific passes, where each task updates its own low-rank matrices, and integrates these updates through global shared-subspace fusion. Specifically, we introduce a shared-subspace resolver based on singular value decomposition (SVD), referred to as the SVD resolver, which extracts a common low-dimensional orthonormal basis across task adapters and projects task updates into this shared space before merging them into a single deployment-time update. The resulting update can be folded into the backbone, preserving the inference structure of the original model without additional runtime branches.
Experiments with Swin Transformer backbones on PASCAL-Context, NYUD-v2, and CelebA show that CRLoMTL achieves a favorable accuracy--parameter trade-off, improving over single-task baselines by up to 2.77\% on PASCAL-Context, 5.03\% on NYUD-v2, and 3.50\% on CelebA, while training only a small fraction of the parameters used by single-task models.


\end{abstract}

\section{Introduction}
\label{sec:intro}
Multi-task learning (MTL) trains a single model to address multiple tasks by
sharing intermediate representations and parameters. This paradigm improves
sample efficiency through cross-task knowledge transfer, reduces the cost of
training separate task-specific models, and lowers memory use and FLOPs by
reusing a common backbone~\cite{vandenhende2021multi,zhang2021survey}.
MTL has shown strong utility across computer vision and natural language
processing, where multiple related objectives can benefit from shared
representations~\cite{ruder2017overview,collobert2008unified}.

Despite these advantages, MTL often suffers from task interference or negative
transfer, where optimizing one task degrades the performance of another
~\cite{wang2020gradient,yu2020gradient}. A common source of interference is
gradient conflict, which occurs when task gradients point in opposing directions.
Such conflicts can slow convergence, destabilize training, and bias the shared
model toward tasks with dominant or easier objectives. Existing methods mitigate
this issue through loss re-weighting~\cite{kendall2018multi}, gradient projection
or surgery~\cite{wang2020gradient,yu2020gradient,zhu2023prompt}, and adaptive
multi-objective optimization~\cite{sener2018multi,liu2021conflict,liu2023famo}.
However, these approaches often require task-wise gradient computation,
optimization subproblems, or additional training-time mechanisms.

Low-rank adaptation provides an alternative way to model task-specific updates
efficiently in parameter space. Instead of directly manipulating task gradients,
low-rank matrices can isolate task-specific adaptation directions while keeping
the shared backbone frozen. This motivates CRLoMTL, a parameter-efficient MTL
framework that mitigates task interference by learning task-specific low-rank
updates and fusing them into a unified model.

Our contributions are as follows:
\begin{itemize}
    \item We introduce a multi-pass training strategy that alternates between
    task-specific passes and global passes. Task-specific passes learn separate
    low-rank adaptation matrices for each task, while global passes construct a
    fused update by resolving task-specific adapters through a common
    low-rank subspace.

    \item Unlike gradient-manipulation methods, CRLoMTL addresses negative transfer
    through low-rank parameter-space adaptation, producing a single fused update
    that can be merged into the backbone for inference without task-specific
    adapter selection or runtime fusion.

    \item We devise a shared-subspace fusion rule that extracts a low-dimensional
    orthonormal basis common to task-specific LoRA matrices and synthesizes a
    fused update through a closed-form least-squares projection.

    \item We evaluate CRLoMTL using Swin Transformer backbones on multiple MTL
    benchmarks, demonstrating a favorable accuracy--parameter trade-off compared
    with existing MTL baselines.
\end{itemize}

CRLoMTL balances task-specific adaptation with global shared-update construction,
improving overall multi-task performance while using substantially fewer
trainable parameters than existing MTL baselines and maintaining practical
training-time and memory overhead.
The rest of the paper is organized as follows. We discuss background and related
work in Section~\ref{sec:background}, introduce CRLoMTL in
Section~\ref{sec:method}, present experiments in Section~\ref{sec:expr}, and
conclude in Section~\ref{sec:conclusion}.

\section{Background \& Related Work}
\label{sec:background}
\subsection{Multi-task Learning \& Challenges}
Multi-task learning (MTL) trains a single model to solve multiple related tasks by
sharing representations, often improving knowledge transfer and reducing the cost
of training separate task-specific models~\cite{vandenhende2021multi,zhang2021survey}.
Given a model $f(x;\theta)$ and $T$ task losses $\{L_t\}_{t=1}^{T}$, the standard
MTL objective minimizes a weighted sum of task losses:
\begin{equation}
    L(\theta) = \sum_{t=1}^T w_t L_t(f(x; \theta)),
\end{equation}
where $w_t$ controls the contribution of task $t$. MTL has been widely used in
computer vision, natural language processing, and healthcare
~\cite{kokkinos2017ubernet,collobert2008unified,moeskops2016deep}.

Despite these benefits, MTL can suffer from negative transfer, where sharing
representations degrades one or more tasks~\cite{wang2019characterizing}. This
can arise from task imbalance, differences in task difficulty or data scale, and
conflicting optimization directions~\cite{chen2018gradnorm}. A common indicator
of task conflict is a negative inner product between task gradients:
\begin{equation}
    \langle \nabla_\theta L_i(\theta), \nabla_\theta L_j(\theta) \rangle < 0,
\label{eq:conflict}
\end{equation}
where $\nabla_\theta L_i(\theta)$ and $\nabla_\theta L_j(\theta)$ denote gradients
from tasks $i$ and $j$, respectively. Such conflicts can slow convergence,
destabilize training, and degrade task performance~\cite{yu2020gradient}. In
addition, MTL performance depends strongly on architectural choices, such as
which layers to share and where to introduce task-specific components
~\cite{ruder2017overview,sener2018multi}.
\subsection{Gradient-Based Methodologies}

{Gradient-based MTL methods aim to reduce negative transfer by modifying task gradients or adaptively weighting task objectives. PCGrad~\cite{yu2020gradient} projects conflicting task gradients when their dot product is negative, while CAGrad~\cite{liu2021conflict} searches for a conflict-averse update direction that balances task improvement with proximity to the average gradient. MGDA~\cite{desideri2012multiple} formulates MTL as multi-objective optimization and computes a common descent direction toward Pareto-optimal solutions. Other methods focus on balancing task learning dynamics: GradNorm~\cite{chen2018gradnorm} adjusts task weights based on relative training rates, DWA~\cite{liu2019end} uses recent loss-change ratios to assign dynamic weights, IMTL-G~\cite{liu2021imtl} balances gradient magnitudes and directions to avoid task dominance, and FAMO~\cite{liu2023famo} provides an efficient adaptive optimization strategy with constant memory and computational overhead.
Recent methods further consider fairness, alignment, and consistency across tasks. Nash-MTL~\cite{navon2022multi} casts MTL as a bargaining game to obtain a balanced shared update direction. FairGrad~\cite{ban2024fairgrad} promotes fair resource allocation among tasks, while Go4Align~\cite{shen2024go4align} improves task alignment through group optimization. ConsMTL~\cite{qin2025consmtl} highlights the importance of task-specific parameters for consistent multi-task optimization. Although these methods mitigate gradient conflict and task imbalance, they often introduce additional computational cost through task-wise gradient computation, optimization subproblems, or auxiliary balancing mechanisms, making efficiency a key challenge for scalable MTL.}

\subsection{Architectural Approaches}
{Beyond gradient-based optimization, several works address MTL challenges through architectural design, aiming to balance shared representations with task-specific adaptation. PAD-Net~\cite{xu2018pad} promotes cross-task information flow by using auxiliary task predictions as intermediate representations for final task prediction. MTI-Net~\cite{vandenhende2020mti} extends this idea with multi-scale feature fusion and task-specific refinement modules, enabling interactions across tasks and scales. Other methods, such as AdaShare~\cite{sun2020adashare}, learn adaptive sharing policies to determine which layers should be shared across tasks. Recently, transformer-based MTL architectures have gained increasing attention~\cite{ye2022inverted,ye2022taskprompter}. InvPT~\cite{ye2022inverted} introduces task-specific inverted bottleneck modules on top of a shared transformer backbone, while MTLoRA~\cite{agiza2024mtlora} leverages low-rank matrices to model task interactions efficiently. RobuMTL~\cite{robumtl} extends this direction to robust MTL by dynamically selecting and fusing perturbation-specific hierarchical LoRA experts under adverse weather conditions. Although these architectures can capture complex task relationships, many require additional parameters, task-specific branches, or extra inference-time modules~\cite{vandenhende2020mti,xu2018pad,agiza2024mtlora}, which may increase computational overhead. Moreover, understanding when MTL is beneficial and how to quantify task relatedness remains an active research direction~\cite{zamir2018taskonomy}.}
\begin{figure*}[h]
\centering
\includegraphics[width=\textwidth]{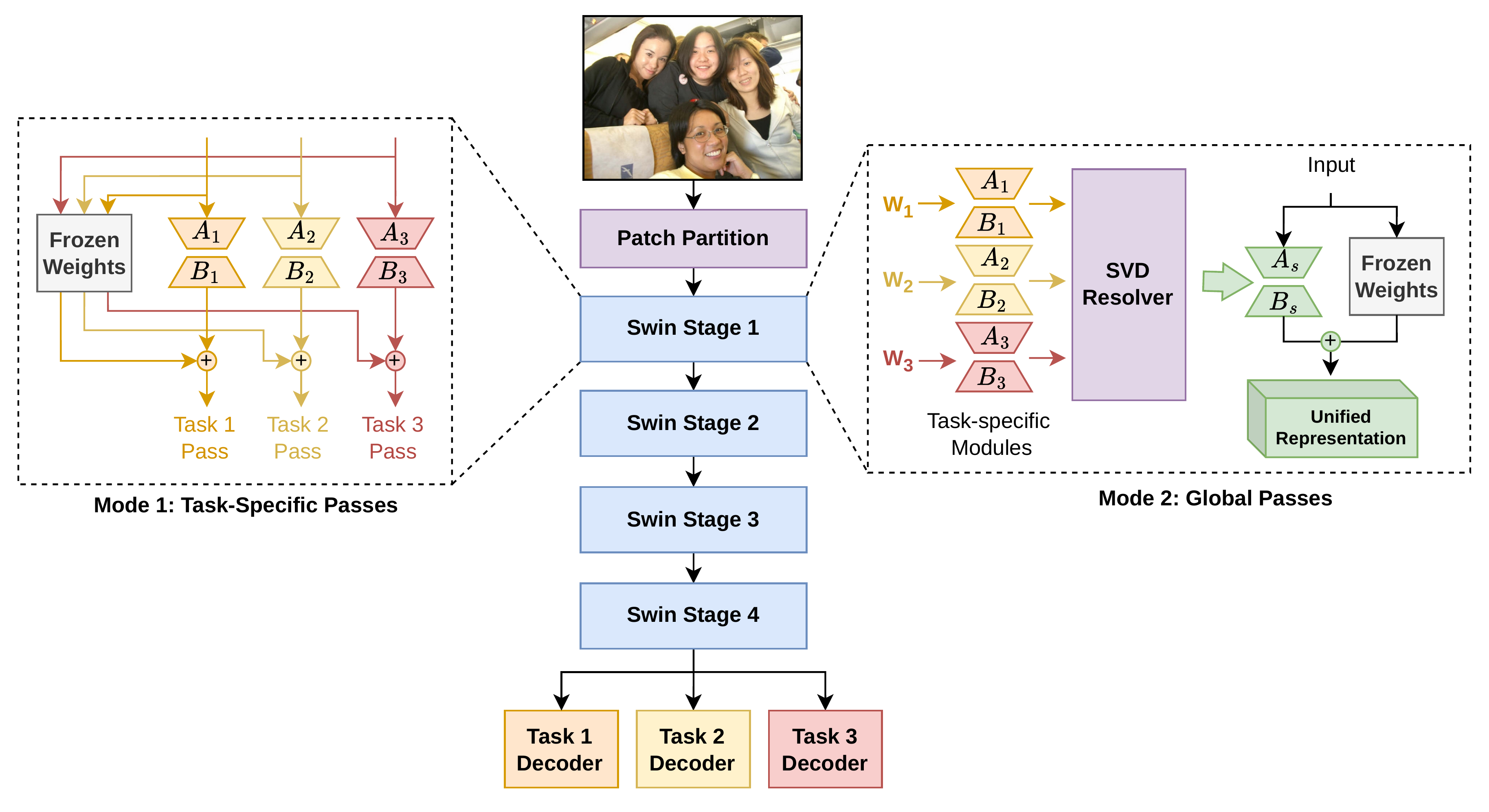}
\vspace{4pt}
\caption{
Overview of the CRLoMTL architecture and training modes. The model alternates between task-specific passes, where each task updates its own low-rank matrices, and global passes, where the SVD resolver fuses task-specific LoRA updates into a single fused adapter \(B_s A_s\). The compact notation \(A_s\) and \(B_s\) denote the shared matrices \(A_{\ell,\mathrm{shared}}\) and \(B_{\ell,\mathrm{shared}}\) produced by the resolver, with the layer index \(\ell\) omitted in the diagram for visual clarity.
}
\label{fig:overview}
\end{figure*}
\subsection{Parameter-Efficient Fine-Tuning (PEFT),  Low-Rank Adaptation (LoRA), and Model Merging}
Recent advancements in Parameter-Efficient Fine-Tuning (PEFT) techniques have shown promise in adapting large pre-trained models to specific tasks with minimal computational overhead. Among these techniques, Low-Rank Adaptation (LoRA) \cite{hu2021lora,dettmers2024qlora} has gained significant attention due to its simplicity and effectiveness.
LoRA introduces task-specific low-rank matrices to adapt pre-trained model weights. For a pre-trained weight matrix $W \in \mathbb{R}^{d \times k}$, LoRA defines the adapted weight as:
\begin{equation}
W' = W + BA,
\end{equation}
\noindent where $B \in \mathbb{R}^{d \times r}$ and $A \in \mathbb{R}^{r \times k}$ are low-rank matrices (typically $r \ll \min(d,k)$). During fine-tuning, only $A$ and $B$ are updated, significantly reducing the number of trainable parameters.
PEFT methods like LoRA offer several advantages in terms of efficiency and flexibility. They require far less memory than full fine-tuning, as most model parameters remain frozen. Training is faster due to the reduced number of trainable parameters. Additionally, multiple task-specific adaptations can be stored and switched easily without modifying the base model. These characteristics make PEFT techniques particularly attractive for MTL scenarios, where they can potentially capture task-specific information efficiently while maintaining a shared base model.

\paragraph{LoRA and Model Merging.}Task arithmetic~\cite{ilharco2023taskarithmetic}, TIES-Merging~\cite{yadav2023ties}, and KnOTS~\cite{stoica2024knots} combine independently fine-tuned task models or adapters in parameter space, primarily as post-hoc merging procedures. Their evaluations largely consider experts with compatible prediction interfaces, such as models trained on different classification datasets, whereas CRLoMTL targets heterogeneous dense MTL, where tasks coexist within the same model but differ in output spaces, losses, and evaluation metrics. More importantly, CRLoMTL performs recurrent differentiable fusion during training: after the task-specific updates of each mini-batch, the resolver constructs a fused adapter whose global MTL loss backpropagates to the task adapters. Thus, the key distinction is training-time repeated fusion for heterogeneous MTL rather than one-time post-hoc model merging.
\subsection{Problem Formulation}
In this section, we formally define the MTL problem and introduce the notation used throughout the paper.
Let $\mathcal{D} = \{(x_i, y_i^1, ..., y_i^T)\}_{i=1}^N$ be a dataset of $N$ samples, where $x_i \in \mathcal{X}$ represents the input features and $y_i^t \in \mathcal{Y}^t$ is the corresponding label for task $t \in \{1, ..., T\}$. We consider a shared model $f_\theta: \mathcal{X} \rightarrow \mathcal{Y}^1 \times ... \times \mathcal{Y}^T$ with parameters $\theta$, which aims to learn $T$ tasks simultaneously.
For each task $t$, we define a task-specific loss function $\mathcal{L}_t: \mathcal{Y}^t \times \mathcal{Y}^t \rightarrow \mathbb{R}$ that measures the discrepancy between the predicted output and the ground truth. The MTL objective can then be formulated as:
\begin{equation}
    \min_\theta \sum_{t=1}^T w_t \frac{1}{N} \sum_{i=1}^N \mathcal{L}_t(f_\theta^t(x_i), y_i^t),
\end{equation}
\noindent where $w_t > 0$ is the weight assigned to task $t$, and $f_\theta^t(x)$ denotes the output of the model for task $t$ given input $x$.
The gradient of the overall loss with respect to the model parameters $\theta$ is given by:

\begin{equation}
    \nabla_\theta \mathcal{L}(\theta) = \sum_{t=1}^T w_t \nabla_\theta \mathcal{L}_t(\theta),
\end{equation}

\noindent where $\nabla_\theta \mathcal{L}_t(\theta)$ is the gradient of the loss for task $t$.
As discussed earlier, gradient conflicts occur when the gradients from different tasks point in opposing directions, as described in Equation \ref{eq:conflict}.
The primary challenge in MTL is finding a set of parameters $\theta$ that performs well across all tasks, even in the presence of conflicting gradients. This involves balancing the potentially competing objectives of different tasks and finding a compromise that leads to good overall performance. Hence, our proposed framework attempts to leverage task-specific low-rank matrices to adapt the model parameters for each task and find an effective shared representation for all the tasks.

\section{Methodology}
\label{sec:method}
CRLoMTL addresses multi-task learning challenges through low-rank matrix adaptations that operate directly on the model's parameter space. As illustrated in Figure~\ref{fig:overview}, our framework employs a multi-pass training mechanism that alternates between task-specific parameter updates and global shared-representation learning. This design enables the simultaneous capture of task-specific features and shared knowledge through three key components: (1) adaptive low-rank decomposition for task specialization, (2) task-specific passes that reduce direct gradient interference, and (3) global shared-subspace fusion that constructs a conflict-aware fused update.
\subsection{Low-Rank Task Adaptation}
For each task $t \in \{1,\ldots,T\}$, we implement parameter adaptation through learnable low-rank matrices:

\begin{equation}
    W_{\ell,t} = W_\ell + {B_{\ell,t} A_{\ell,t}},
     \label{eq:task_lora_adaptation}
\end{equation}
where $W_\ell \in \mathbb{R}^{d \times k}$ denotes a frozen backbone weight matrix at LoRA-equipped layer $\ell$, and
$B_{\ell,t} \in \mathbb{R}^{d \times r}$ and $A_{\ell,t} \in \mathbb{R}^{r \times k}$ are the task-specific low-rank adaptation matrices for task $t$, with $r \ll \min(d,k)$.
These low-rank components are inserted at selected adaptation points,
providing task-specific adaptation while keeping the backbone frozen.
Factorization reduces trainable parameters while retaining sufficient
expressivity for task specialization.
\subsection{Multi-Pass Training with Shared-Subspace Fusion}
CRLoMTL performs alternating optimization within each mini-batch: the $T$ task adapters are updated sequentially and independently during the \emph{task-specific passes}, with gradients cleared between tasks so their gradients are not explicitly combined. The updated adapters are then passed to the SVD resolver to construct a fused adapter, followed by one \emph{global pass} using the aggregate MTL loss. Since the fused adapter remains differentiably connected to the persistent task adapters, the global loss backpropagates through the resolver and updates all task adapters; the adapters are neither overwritten nor reset. Thus, each cycle consists of $T$ task-specific updates followed by one global fused update.

\subsubsection{Task-Specific Passes}
During task-specific passes, each task $t$ independently updates its own layer-wise low-rank matrices 
$\{A_{\ell,t}, B_{\ell,t}\}_{\ell \in \mathcal{L}}$.
In these passes, the forward computation for each task incorporates both the frozen backbone and its task-specific parameters. 
Specifically, for a given task $t$, each LoRA-equipped layer $\ell \in \mathcal{L}$ uses the task-specific adapted weight defined in Eq.~\eqref{eq:task_lora_adaptation} during the forward pass, where $W_\ell$
denotes the frozen backbone weight matrix at layer $\ell$. In the backward pass, only the task-specific LoRA parameters $\{A_{\ell,t}, B_{\ell,t}\}_{\ell \in \mathcal{L}}$ for the current task $t$ are updated. Crucially, the frozen backbone weights $\{W_\ell\}_{\ell \in \mathcal{L}}$
and the low-rank matrices of all other tasks $\{A_{\ell,j}, B_{\ell,j}\}_{\ell \in \mathcal{L}}$ for $j \neq t$ remain unchanged during this process. The adapters of other tasks are not used in the forward pass for task $t$ and are not updated during backpropagation. Gradients are cleared between task-specific passes, preventing explicit accumulation or combination of gradients across tasks.
{This design enables each task to learn task-specific adaptation directions while reducing direct gradient interference, since only the LoRA parameters of the current task are updated during its task-specific pass.}

\subsubsection{Global Passes}
\label{sec:global_pass}

In the global passes (shown in Figure~\ref{fig:overview}), CRLoMTL constructs a unified representation from the task-specific low-rank adapters using shared-subspace fusion. Given the current task-specific adapters $\{A_{\ell,t}, B_{\ell,t}\}_{t=1}^{T}$ at each LoRA-equipped layer $\ell$, the global pass computes a shared basis using the resolver described in Section~\ref{sec:shared_resolver}. Each task update is then projected onto this basis and aggregated to obtain a single fused low-rank update:
\begin{equation}
W_{\ell,\mathrm{unified}}
=
W_\ell
+
\mathbf{B}_{\ell,\mathrm{shared}}
\mathbf{A}_{\ell,\mathrm{shared}}.
\label{eq:global_svd_fusion}
\end{equation}

After all $T$ task-specific updates in the current mini-batch, the fused adapter is used for one task-agnostic forward pass with the aggregate MTL loss. Because the fused update remains differentiably connected to the persistent task adapters, the global backward pass propagates through the resolver and updates all task-specific adapters; they are neither overwritten nor reset.

The global pass acts as a conflict-aware consolidation step. While task-specific passes allow each task to learn specialized adaptation directions, the global pass merges these directions through a common low-dimensional subspace. This encourages the fused update to retain directions shared across tasks while reducing the influence of residual task-specific or conflicting components. After training, the resulting low-rank update is folded into the corresponding backbone weights, so the deployed model uses a single merged weight matrix without additional adapter branches or runtime fusion operations.

\subsection{SVD-Based Shared-Subspace Resolver} \label{sec:shared_resolver}
The global pass in Section~\ref{sec:global_pass} requires a structured fusion mechanism for constructing a fused update from task-specific low-rank adapters. Directly averaging task-specific updates in the original parameter space may preserve task-specific or conflicting directions. We therefore introduce the \emph{SVD resolver}, which uses truncated SVD to identify a common subspace for all tasks and then projects updates onto this basis via a least-squares solution.
Let 
$\{A_{\ell,t}\in\mathbb{R}^{r\times d_{\ell}^{\mathrm{in}}},
B_{\ell,t}\in\mathbb{R}^{d_{\ell}^{\mathrm{out}}\times r}\}_{t=1}^{T}$
be the task-specific low-rank adapters learned at a LoRA-equipped layer
$\ell \in \mathcal{L}$. 
The resolver is applied independently at each LoRA-equipped layer. We use
$A_{\ell,t}$ for the shared input-side subspace of
$\Delta W_{\ell,t}=B_{\ell,t}A_{\ell,t}$, while $B_{\ell,t}$ retains
task-dependent output mappings; $B$-based extraction reduces the average relative
improvement metric $\Delta_m$ (defined in Eq.~\eqref{metric}) by $0.03$
percentage points.
We concatenate the transposed $A$ matrices as
\begin{equation}
\mathbf{A}_{\ell,\mathrm{cat}}
=
\bigl[
A_{\ell,1}^{\top}
\;|\;
A_{\ell,2}^{\top}
\;|\;
\dots
\;|\;
A_{\ell,T}^{\top}
\bigr]
\in
\mathbb{R}^{d_{\ell}^{\mathrm{in}}\times Tr}.
\label{eq:svd_concat}
\end{equation}
We use the learned $A_{\ell,t}$ directly without additional normalization; Frobenius normalization changes $\Delta_m$ by only $0.001$ percentage points, indicating negligible sensitivity to adapter magnitude.
\paragraph{Basis extraction.}
Applying the truncated SVD
$\mathbf{A}_{\ell,\mathrm{cat}}
=
\mathbf{U}_{\ell}\boldsymbol{\Sigma}_{\ell}\mathbf{V}_{\ell}^{\top}$
and retaining the leading $r_s$ left singular vectors yields
\begin{equation}
\begin{aligned}
\mathbf{U}_{\ell,s}
&= \mathbf{U}_{\ell}[:,1:r_s], \\
\mathbf{A}_{\ell,\mathrm{shared}}
&= \mathbf{U}_{\ell,s}^{\top}
\in \mathbb{R}^{r_s\times d_{\ell}^{\mathrm{in}}}, \\
r_s
&= \min(\max(1,\lfloor r/2\rfloor),\,d_{\ell}^{\mathrm{in}}).
\end{aligned}
\label{eq:shared_basis}
\end{equation}
Since $\mathbf{U}_{\ell,s}$ has orthonormal columns, 
$\mathbf{A}_{\ell,\mathrm{shared}}$ has orthonormal rows.

\paragraph{Projection and aggregation.}
For each task, we first define the task-specific LoRA update at layer $\ell$ as
\begin{equation}
\Delta W_{\ell,t}=B_{\ell,t}A_{\ell,t}.
\label{eq:lora_update}
\end{equation}
We then project this update onto the shared basis using least squares:
\begin{equation}
\mathbf{X}_{\ell,t}
=
\underset{\mathbf{X}\in\mathbb{R}^{d_{\ell}^{\mathrm{out}}\times r_s}}{\arg\min}
\;
\left\|
\Delta W_{\ell,t}
-
\mathbf{X}\mathbf{A}_{\ell,\mathrm{shared}}
\right\|_F^2
=
\Delta W_{\ell,t}\mathbf{A}_{\ell,\mathrm{shared}}^{\top},
\label{eq:ls_projection}
\end{equation}
where the closed form follows from the orthonormality of
$\mathbf{A}_{\ell,\mathrm{shared}}$, i.e.,
$\mathbf{A}_{\ell,\mathrm{shared}}
\mathbf{A}_{\ell,\mathrm{shared}}^{\top}
=
\mathbf{I}$.
The fused adapter at layer $\ell$ is then

\begin{equation}
\begin{aligned}
\mathbf{B}_{\ell,\mathrm{shared}}
&= \frac{1}{T}\sum_{t=1}^{T}\mathbf{X}_{\ell,t}, \\
W_{\ell,\mathrm{unified}}
&= W_\ell +
\mathbf{B}_{\ell,\mathrm{shared}}\mathbf{A}_{\ell,\mathrm{shared}}.
\end{aligned}
\label{eq:shared_fusion}
\end{equation}
The global loss backpropagates through the resolver to all task-specific LoRA parameters. After training, the fused update is folded into the backbone, adding no inference overhead. The complete shared-subspace resolution procedure is summarized in Algorithm \ref{alg:ssr}.
\begin{algorithm}[t]
\caption{SVD Resolver at Layer $\ell$}
\label{alg:ssr}
\begin{algorithmic}[1]
\Procedure{SVDResolver}{$\{A_{\ell,t},B_{\ell,t}\}_{t=1}^{T}, r_s$}
  \State $\mathbf{A}_{\ell,\mathrm{cat}}
  \leftarrow
  \mathrm{cat}(\{A_{\ell,t}^{\top}\}_{t=1}^{T},\;\mathrm{dim}=1)$

  \State $(\mathbf{U}_{\ell}, \boldsymbol{\Sigma}_{\ell}, \mathbf{V}_{\ell})
  \leftarrow
  \mathrm{svd}(\mathbf{A}_{\ell,\mathrm{cat}},\;\mathrm{full\_matrices}=\mathrm{False})$

  \State $\mathbf{U}_{\ell,s} \leftarrow \mathbf{U}_{\ell}[:, :r_s]$
  \State $\mathbf{A}_{\ell,\mathrm{shared}} \leftarrow \mathbf{U}_{\ell,s}^{\top}$

  \State $\mathbf{B}_{\mathrm{acc}}
  \leftarrow
  \mathbf{0}_{d_{\ell}^{\mathrm{out}}\times r_s}$

  \For{$t=1$ to $T$}
    \State $\Delta W_{\ell,t} \leftarrow B_{\ell,t}A_{\ell,t}$
    \State $\mathbf{X}_{\ell,t}
    \leftarrow
    \Delta W_{\ell,t}\mathbf{A}_{\ell,\mathrm{shared}}^{\top}$
    \State $\mathbf{B}_{\mathrm{acc}} \mathrel{+}= \mathbf{X}_{\ell,t}$
  \EndFor

  \State $\mathbf{B}_{\ell,\mathrm{shared}}
  \leftarrow
  \mathbf{B}_{\mathrm{acc}}/T$

  \State $\Delta W_{\ell,\mathrm{fused}}
  \leftarrow
  \mathbf{B}_{\ell,\mathrm{shared}}\mathbf{A}_{\ell,\mathrm{shared}}$
  \State \Return $\Delta W_{\ell,\mathrm{fused}}$
\EndProcedure
\end{algorithmic}
\end{algorithm}

\section{Experimental Results}
\label{sec:expr}
\renewcommand{\arraystretch}{1.06}

\subsection{Setup}
We implement CRLoMTL in PyTorch and will release the code and pretrained weights upon publication.
All experiments use a Swin Transformer backbone~\cite{liu2021swin},
pre-trained on ImageNet-1K~\cite{deng2009imagenet}, as the shared encoder.
For dense prediction, we attach simple task-specific decoders for different dense tasks, similar to the one in HR-Net \cite{wang2020deep}, which includes linear and bilinear upsampling layers. We use LoRA rank $r=8$ with scaling $\alpha/r=2$ ($\alpha=16$).  
We evaluate our method primarily on two standard multi-task scene understanding
benchmarks: PASCAL-Context~\cite{pascal} and NYUD-v2~\cite{nyud}. Following
common practice in MTL literature~\cite{vandenhende2020mti,liu2019end}, we use
the standard PASCAL-Context split of 4,998 training and 5,105 validation images,
with annotations for semantic segmentation, human-part segmentation, surface
normal estimation, and saliency detection. For NYUD-v2, we use the standard split
of 795 training and 654 testing images and evaluate semantic segmentation and
depth estimation. For the dense prediction benchmarks, decoder parameters account for only $7\%$ of the total trainable parameters when Swin-Tiny is used as the backbone.

In addition, we evaluate CRLoMTL on CelebA~\cite{liu2015faceattributes}, an
image-level multi-attribute classification benchmark with over 200K face images
and 40 binary facial attributes, which we treat as 40 binary classification
tasks. This experiment assesses whether CRLoMTL extends beyond dense scene
understanding to many-task classification.
To train our MTL model, we use a weighted sum of task losses:
\begin{equation}
\text{Loss}_{\text{MTL}} = \sum_{i=1}^T \omega_{\text{task } i} L_{\text{task } i},
\end{equation}
where $\omega_{\text{task } i}$ and $L_{\text{task } i}$ denote the weight and loss for task $i$, respectively. For PASCAL-Context and NYUD-v2, we use pixel-wise cross-entropy for semantic and human-part segmentation, balanced binary cross-entropy for saliency, and $L_1$ loss for surface normals and depth estimation. We adopt the task weights used by Vandenhende \textit{et al.}~\cite{vandenhende2020mti}. For CelebA, we use binary cross-entropy over the 40 image-level facial attribute classification tasks.
{
Accuracy experiments were conducted on a V100, while efficiency profiling was conducted on an H100 under a controlled setup using batch size 32. 
For the dense prediction experiments, all methods use AdamW with cosine decay, weight decay $0.05$, batch size $32$, and $500$ epochs, with $448\times448$ inputs for dense prediction and $224\times224$  for CelebA.
}
\begin{table*}[t]
\centering
\caption{Evaluation of CRLoMTL versus state-of-the-art (SOTA) methods on PASCAL-Context dataset using Swin-Tiny backbone. $\Delta_m$ represents the relative improvement of the MTL model compared to the single-task models.
}
\label{tab:eval_tiny}
\vspace{6pt}
\setlength{\tabcolsep}{3.5pt} 
\begin{tabular}{lcccccc}
\hline
Method & SemSeg & Human Parts & Saliency & Normals & Trainable & $\Delta_m$(\%) $\uparrow$  \\
       & (mIoU $\uparrow$) & (mIoU $\uparrow$) & (mIoU $\uparrow$) & (RMSE $\downarrow$) & Parameters (M) & \\
\hline
Single Task & 67.21 & 61.93 & 62.35 & 17.97  & 112.62 & 0\\
Vanilla MTL  & 67.56 & 60.24 & 65.21 & 16.64 & 29.47 & +2.23 \\
\hline
MTI-Net & 66.93 & 59.66 & 64.76 & 16.23 & 31.47 & +2.37 \\
MTLoRA & 67.90 & 59.84 & 65.40 & 16.60 & 8.34 & +2.55 \\
\hline
{KnOTS} &67.90&60.26&64.68&16.81&30.29&+2.12
\\
\hline
{DWA} &66.88&60.0 &64.86  &16.40& 29.47 &+2.29 \\ 
PCGrad & 67.19 & 60.50 & 64.90 & 16.52 & 29.47 & +2.46 \\
CAGrad & 67.53 & 60.59 & 64.87 & 16.58 & 29.47 & +2.51 \\
NashMTL & 67.25 & 60.60 & 64.82 & 16.68 & 29.47 & +2.26 \\
MGDA & 66.21 & 60.12 & 64.92 & 16.31 & 29.47 & +2.22 \\

{IMTL-G} & 67.49 & 60.80 & 65.16 &16.84& 29.47 &  +2.34 \\ 
{FAMO} & 68.45 & 61.23 & 64.96 &17.79& 29.47 & +1.47 \\ 
{FairGrad} & 67.99&60.76&64.82  & 17.03& 29.47 &+2.11 \\ 
{Go4Align} &67.96 &60.91  &65.01 &17.55 & 29.47 &+1.51 \\ 
{ConsMTL} &66.95 &60.00  & 65.46 &16.89 & 29.47 &+1.87 \\ \hline
\textbf{CRLoMTL} &  {67.86} & {60.43} & {64.93} & {16.48} & \textbf{4.39} & {\textbf{+2.77}} \\
\hline
\end{tabular}
\end{table*}
\vspace{-10pt}
\renewcommand{\arraystretch}{1.06}
\begin{table*}[h]
\centering
\caption{Evaluation on PASCAL-Context with a Swin-Small backbone.}
\label{tab:eval_small}
\vspace{6pt}
\setlength{\tabcolsep}{3.5pt} 
\begin{tabular}{lcccccc}
\hline
Method & SemSeg & Human Parts & Saliency & Normals & Trainable & $\Delta_m$(\%) $\uparrow$ \\
       & (mIoU $\uparrow$) & (mIoU $\uparrow$) & (mIoU $\uparrow$) & (RMSE $\downarrow$) & Parameters (M) & \\
\hline
Vanilla MTL  & 70.20 & 63.30 & 66.09 & 16.17 & 50.78 &  +5.66 \\
\textbf{CRLoMTL} & 70.78 & 63.60 & 66.28 & 16.17 & \textbf{6.75} & \textbf{+6.08} \\

\hline
\end{tabular}
\end{table*}

\subsection{Baseline Methods}
We compare CRLoMTL with representative state-of-the-art (SOTA) MTL methods from five categories: task-interaction architectures, parameter-efficient adaptation, dynamic loss weighting, gradient-balancing optimization, and parameter-space merging controls. These include MTI-Net~\cite{vandenhende2020mti}, MTLoRA~\cite{agiza2024mtlora}, DWA~\cite{liu2019end}, PCGrad~\cite{yu2020gradient}, CAGrad~\cite{liu2021conflict}, Nash-MTL~\cite{navon2022multi}, MGDA~\cite{desideri2012multiple}, IMTL-G~\cite{liu2021imtl}, FAMO~\cite{liu2023famo}, FairGrad~\cite{ban2024fairgrad}, Go4Align~\cite{shen2024go4align}, and ConsMTL~\cite{qin2025consmtl}. We additionally include Shared-LoRA and a KnOTS-style SVD merging control~\cite{stoica2024knots} to isolate the benefit of CRLoMTL’s recurrent training-time fusion over shared adaptation and post-hoc-style LoRA merging. All methods are implemented within the same experimental framework using the same backbone and datasets, with only the MTL or adapter-fusion strategy varied.






\subsection{Evaluation Results on PASCAL-Context}

We use mean intersection over union (mIoU) to evaluate semantic segmentation, saliency estimation, and human part segmentation. For the surface normal task, we use the root-mean-square error (RMSE) in the predicted angles. To measure overall performance, we use $\Delta_m$, which is the average per-task improvement in performance compared to the single-task baseline:

\begin{equation}
\Delta_m = \frac{1}{T} \sum_{i=1}^T (-1)^{l_i} \frac{(M_i - M_{st,i})}{M_{st,i}},
\label{metric}
\end{equation}
where $l_i = 1$ if a lower value means better for performance measure $M_i$ of task $i$, and 0 otherwise. Single-task performance is measured for a fully converged model using the same backbone network only for that task.
Tables \ref{tab:eval_tiny} and \ref{tab:eval_small} present the results for models using Swin-Tiny and Swin-Small backbones, respectively. We compare CRLoMTL against existing SOTA MTL methods and single-task baselines. {The reported parameter counts refer to trainable parameters during training.}
As shown in Table \ref{tab:eval_tiny}, CRLoMTL improves the single-task baseline by $2.77\%$ while training only 4\% of the parameters when using a Swin-Tiny backbone, demonstrating a favorable trade-off between accuracy and parameter efficiency.
On the larger Swin-Small backbone (Table 2), CRLoMTL delivers a 6.08\% gain while training less than 6\% of the parameters, demonstrating that its effectiveness is maintained as model capacity increases.
Additionally, unlike methods that retain task-specific branches or adapters at inference, such as MTI-Net~\cite{vandenhende2020mti} and MTLoRA~\cite{agiza2024mtlora}, CRLoMTL folds the fused low-rank update into the backbone after training, yielding a single deployed model without task-specific adapter branches.
The default single-seed result reported in Table~\ref{tab:eval_tiny} is
$\Delta_m = 2.77\%$. Across three random seeds, CRLoMTL obtains
$\Delta_m = 2.74 \pm 0.03\%$; detailed per-task results are provided in the
supplementary material.

\subsubsection{{Analysis of Global Pass}}

\begin{figure}[t]
    \centering
    \includegraphics[width=0.85\linewidth,trim=8 6 8 6,clip]{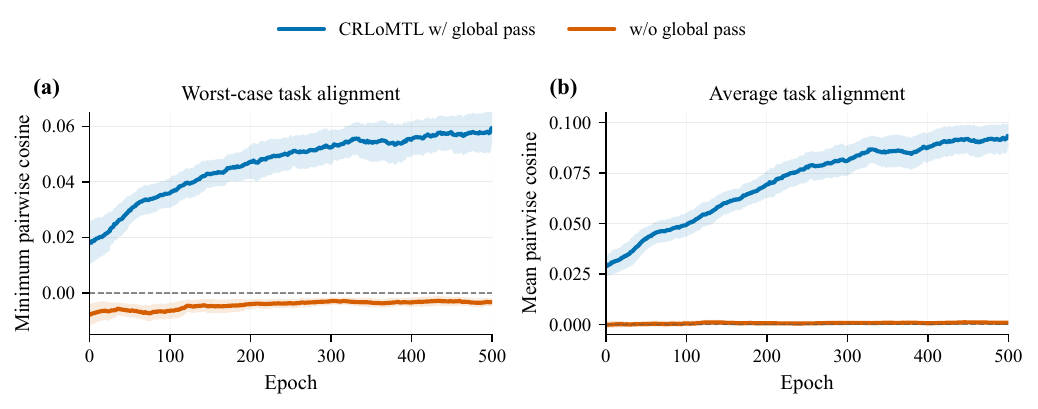}
    \vspace{6pt}
    \caption{Effect of the global pass on task-specific LoRA update alignment.}
    \label{fig:global_pass_alignment}
\end{figure}

To assess the effect of the global pass, we analyze update alignment in the
low-rank parameter space. Using the task-specific LoRA updates
$\Delta W_{\ell,t}$ defined in Eq.~\eqref{eq:lora_update}, we measure their
change over one training cycle as
\begin{equation}
\delta_{\ell,t}
=
\Delta W_{\ell,t}^{(\mathrm{after})}
-
\Delta W_{\ell,t}^{(\mathrm{before})}.
\end{equation}
For full CRLoMTL, ``after'' is measured after the task-specific and global passes;
for the no-global-pass ablation, it is measured after the task-specific passes
only. We flatten the analyzed update changes into one vector per task and compute
pairwise cosine similarity:
\begin{equation}
C_{ij} =
\frac{\langle \delta_i, \delta_j \rangle}
{\|\delta_i\|_2 \|\delta_j\|_2}.
\end{equation}
We report the mean and minimum off-diagonal entries of $C$ as average and
worst-case task-update alignment. Figure~\ref{fig:global_pass_alignment} shows that the global pass increases both
alignment metrics, whereas the no-global-pass ablation remains near zero or
slightly negative. This suggests that the global pass encourages more compatible
task-specific LoRA updates before fusion, without using an explicit alignment loss.

\subsubsection{Analysis of SVD Conflict Suppression}\label{sec:svd_conflict_analysis}
We analyze the SVD resolver in parameter-update space to assess how
the SVD projection affects cross-task compatibility. For each LoRA-equipped layer
$\ell$ and task $t$, we use the task-specific LoRA update
$\Delta W_{\ell,t}$ defined in Eq.~\eqref{eq:lora_update}. Following the
projection in Eq.~\eqref{eq:ls_projection}, we decompose each update into a
component retained in the shared subspace and an orthogonal residual:
\begin{equation}
    \mathbf{K}_{\ell,t}
    =
    \mathbf{X}_{\ell,t}\mathbf{A}_{\ell,\mathrm{shared}},
    \qquad
    \mathbf{R}_{\ell,t}
    =
    \Delta W_{\ell,t}
    -
    \mathbf{K}_{\ell,t}.
    \label{eq:kept_residual}
\end{equation}
The retained component $\mathbf{K}_{\ell,t}$ captures the portion of the task
update represented by the SVD-derived shared basis at layer $\ell$, whereas
$\mathbf{R}_{\ell,t}$ corresponds to the component discarded by the projection.
We quantify cross-task compatibility using the average pairwise cosine similarity
and the negative-pair rate (NPR), defined as the fraction of task pairs with
negative cosine similarity, over the sets
$\{\Delta W_{\ell,t}\}$, $\{\mathbf{K}_{\ell,t}\}$, and
$\{\mathbf{R}_{\ell,t}\}$.

\begin{figure*}[t]
\centering
\includegraphics[width=\linewidth]{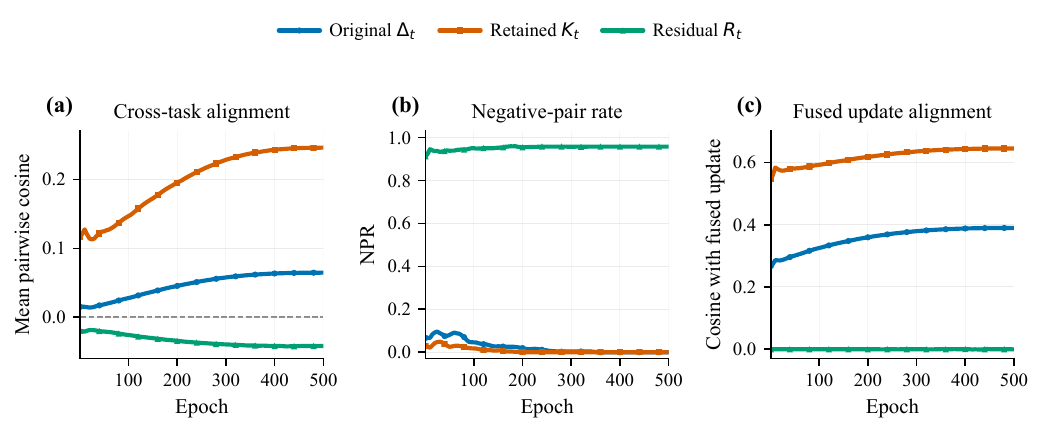}
\vspace{4pt}
\caption{
{SVD resolver analysis in parameter-update space.
(a) The retained component $K_t$ has higher cross-task alignment than the original update $\Delta_t$ and residual $R_t$.
(b) The retained component has lower negative-pair rate (NPR), while the residual contains most conflict-prone directions.
(c) The final fused LoRA update is more aligned with $K_t$ than with $R_t$, indicating that the merge follows the shared SVD-retained subspace. For clarity, the layer index $\ell$ is omitted and $\Delta_t$ abbreviates $\Delta W_{\ell,t}$.
}}
\label{fig:svd_resolver_analysis}
\end{figure*}

As shown in Figure~\ref{fig:svd_resolver_analysis}(a), the retained components
$\mathbf{K}_{\ell,t}$ consistently achieve higher pairwise cosine similarity than
both the original updates $\Delta W_{\ell,t}$ and the residuals
$\mathbf{R}_{\ell,t}$, indicating stronger alignment within the recovered shared
subspace. Figure~\ref{fig:svd_resolver_analysis}(b) further shows that
$\mathbf{K}_{\ell,t}$ has a substantially lower NPR, while the residuals exhibit
the highest NPR. This suggests that the SVD projection concentrates
cross-task-compatible directions in the retained component and relegates less
compatible, task-specific directions to the residual. Finally,
Figure~\ref{fig:svd_resolver_analysis}(c) shows that the fused LoRA update is
more strongly aligned with $\mathbf{K}_{\ell,t}$ than with
$\mathbf{R}_{\ell,t}$, indicating that the final fused adapter is primarily
guided by the shared SVD-retained subspace. Additional retained/residual energy
analysis is provided in the supplementary material.

\vspace{-9pt}
\subsubsection{Efficiency Analysis}
\begin{figure}[t]
    \centering
    \includegraphics[width=\linewidth]{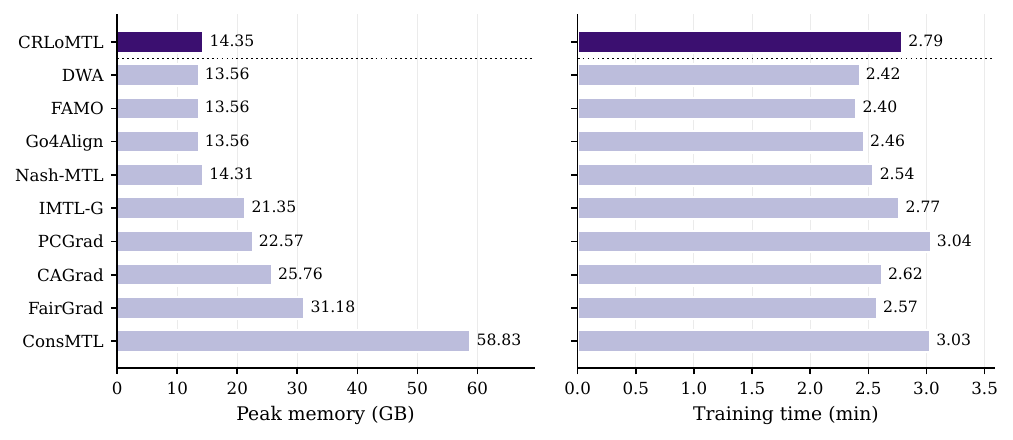}
    \vspace{4pt}
    \caption{{Memory usage and average training time per epoch. 
    }}
    \label{fig:resource_comparison}
\end{figure}

{
}
For efficiency analysis, we evaluate all methods on the same NVIDIA H100 GPU under
an identical PASCAL setup. As shown in Figure~\ref{fig:resource_comparison}, CRLoMTL has a
14.35~GB memory footprint, within 6\% of the lowest baseline footprint (13.56~GB), while
benefiting from its parameter-efficient low-rank updates. Its average training time is 2.79~min/epoch, with
most of the overhead coming from the alternating multi-pass procedure: task-specific passes
account for 46.22\% of the step time, the global backward pass and gradient scaling account
for 33.30\%, and global fusion accounts for only 20.05\%. Thus, the main training cost comes
from multi-pass optimization rather than the SVD-based fusion itself.

As shown in Table~\ref{tab:deployment_analysis}, merging reduces CRLoMTL from 31.91M to 29.47M parameters, matching vanilla MTL, with comparable evaluation time ($229.23\pm2.64$s vs.\ $230.63\pm2.40$s). Before merging, only 4.39M parameters are trainable (2.44M encoder LoRA parameters and 1.95M task-specific decoder parameters), while the folded adapters introduce no inference-time branches or runtime fusion.
\vspace{-9pt}
\begin{table}[h]
\centering
\caption{Deployment analysis on PASCAL-Context over three evaluation runs.}
\label{tab:deployment_analysis}
\vspace{6pt}
{%
\begin{tabular}{lccc}
\hline
Model & Parameters (M) & Eval. Time (s) & Time / Image (ms) \\
\hline
CRLoMTL unmerged & 31.91 & $235.81 \pm 1.57$ & $46.19 \pm 0.31$ \\
CRLoMTL merged & 29.47 & $229.23 \pm 2.64$ & $44.90 \pm 0.52$ \\
Vanilla MTL  & 29.47 & $230.63 \pm 2.40$ & $45.18 \pm 0.47$ \\
\hline
\end{tabular}
}
\end{table}
\begin{table*}[h]
\centering

\begin{minipage}[t]{0.56\linewidth}
\centering
\captionof{table}{Evaluation on NYUD-v2.}
\label{tab:nyud}
\vspace{3pt}

\small
\setlength{\tabcolsep}{2.8pt}
\renewcommand{\arraystretch}{0.93}
\begin{tabular}{lcccc}
\hline
Method & SemSeg & Depth & Trainable & $\Delta_m$ (\%) $\uparrow$\\
       & (mIoU $\uparrow$) & (RMSE $\downarrow$) & Params. (M) & \\
\hline
Single Task & 41.38 & 0.696 & 56.04 & 0 \\
Vanilla MTL & 41.43 & 0.642 & 28.50 & +3.92 \\
DWA         & 41.76 & 0.649 & 28.50 & +3.84 \\
Nash-MTL     & 41.69 & 0.652 & 28.50 & +3.53 \\
PCGrad      & 41.62 & 0.660 & 28.50 & +2.87 \\
CAGrad      & 41.70 & 0.661 & 28.50 & +2.97 \\
IMTL-G      & 41.79 & 0.650 & 28.50 & +3.76 \\
FAMO        & 41.63 & 0.655 & 28.50 & +3.25 \\
FairGrad    & 41.79 & 0.647 & 28.50 & +4.02 \\
Go4Align    & 41.63 & 0.650 & 28.50 & +3.61 \\
ConsMTL      & 41.88 & 0.645 & 28.50 & +4.27 \\
\textbf{CRLoMTL} & 41.65 & 0.630 & \textbf{2.22} & \textbf{+5.03} \\
\hline
\end{tabular}
\end{minipage}
\hfill
\begin{minipage}[t]{0.40\linewidth}
\centering
\captionof{table}{Results on CelebA.}
\label{tab:celeba_results}
\vspace{3pt}

\small
\setlength{\tabcolsep}{6pt}
\renewcommand{\arraystretch}{1.34}
\begin{tabular}{lcc}
\hline
Method & $\Delta_m \uparrow$ & MR $\downarrow$ \\
\hline
STL      & 0.00   & 4.97 \\
FAMO     & -10.51 & 7.15 \\
ConsMTL  & -2.64  & 5.28 \\
FairGrad & -4.21  & 5.38 \\
IMTL-G   & -4.02  & 5.62 \\
Go4Align & +1.24  & 3.10 \\
DWA      & +3.01  & 2.46 \\
\textbf{CRLoMTL} & \textbf{+3.50} & \textbf{2.03} \\
\hline
\end{tabular}
\end{minipage}

\end{table*}
\subsection{Evaluation Results on NYUD-v2}
We evaluate CRLoMTL on NYUD-v2~\cite{nyud}, an indoor scene understanding benchmark with semantic segmentation and depth estimation, using mIoU and RMSE, respectively. As shown in Table~\ref{tab:nyud}, CRLoMTL improves the single-task baseline by $5.03\%$ with only 2.22M trainable parameters, corresponding to 3.96\% of the single-task and 7.79\% of the standard MTL parameter counts.

\subsection{Additional Evaluation on CelebA}
\label{sec:celeba}

We further evaluate CRLoMTL on CelebA~\cite{liu2015faceattributes}, a many-task image-level classification benchmark with 40 tasks. To improve scalability, we retain 40 task-specific classification heads but share LoRA adapters within 7 manually defined semantic groups: hair, facial hair, expression, makeup, accessories, face shape, and identity/style. The grouping is determined solely by attribute semantics.
\paragraph{Grouping and scalability.}
Using 40 per-attribute adapters improves $\Delta_m$ from $+3.50\%$ to
$+4.31\%$ but increases task passes and resolver inputs. In this setting, SVD
accounts for only $8.18\%$ of epoch time, versus $45.43\%$ for task-specific
passes and $34.62\%$ for the global backward.

\noindent\textbf{Setup.}
Following ConsMTL~\cite{qin2025consmtl}, we train for 15 epochs and report MR
and $\Delta_m$ from Eq.~\ref{metric}, where lower MR and higher $\Delta_m$ are
better. CRLoMTL achieves the best overall result in Table~\ref{tab:celeba_results}.
\subsection{Ablation Study}

\noindent\textbf{Shared Subspace Rank Sensitivity:}
Table~\ref{tab:ablations}(a) shows negligible sensitivity to $r_s$, with only $0.006$ percentage-point variation; we therefore use $r_s=r/2$ as a compact default.

\noindent{\textbf{CRLoMTL Components:}}
Table~\ref{tab:ablations}(b) isolates task-specific adaptation and shared-subspace fusion. \textit{Shared-LoRA}, a standard MTL model with one LoRA shared across tasks, shows that the gain is not due to LoRA regularization or backbone freezing alone. The SVD resolver further improves over average fusion with only training-time overhead and no inference cost after merging.

\noindent\textbf{Hyperparameter Sensitivity:}
On PASCAL-Context, we vary LoRA rank with $\alpha=2r$ and $\alpha$ with fixed $r=8$ (Table~\ref{tab:ablations}(c)--(d)). The small variation in $\Delta_m$ across tested ranks and nonzero $\alpha$ values indicates limited sensitivity to these hyperparameters.

\begin{table*}[h]
\centering
\caption{Component ablation and hyperparameter sensitivity on PASCAL-Context.}
\label{tab:ablations}
\vspace{3pt}

\begin{minipage}[t]{0.40\linewidth}
\centering

\textbf{(a) Shared rank $r_s$}\label{tab:ablations_shared}\\[-1pt]
\small
\setlength{\tabcolsep}{5pt}
\renewcommand{\arraystretch}{0.92}
\begin{tabular}{lccc}
\hline
$r_s$ & $r/4$ & $r/2$ & $r$ \\
\hline
$\Delta_m$ (\%) & +2.764 & +2.770 & +2.767 \\
\hline
\end{tabular}

\vspace{4pt}

\textbf{(b) Components}\label{tab:ablations_components}\\[-1pt]
\setlength{\tabcolsep}{5pt}
\begin{tabular}{lc}
\hline
Variant & $\Delta_m$ (\%) \\
\hline
Shared-LoRA & -0.60 \\
Avg. fusion & +2.50 \\
SVD resolver & \textbf{+2.77} \\
\hline
\end{tabular}

\end{minipage}
\hfill
\begin{minipage}[t]{0.57\linewidth}
\centering

\textbf{(c) LoRA rank $r$}\label{tab:rank_ablation}\\[-1pt]
\small
\setlength{\tabcolsep}{5pt}
\renewcommand{\arraystretch}{0.92}
\begin{tabular}{lccccc}
\hline
$r$ & 4 & 8 & 16 & 32 & 64 \\
\hline
$\Delta_m$ (\%) & +2.74 & +2.77 & +2.73 & +2.76 & +2.68 \\
\hline
\end{tabular}

\vspace{4pt}

\textbf{(d) Scaling $\alpha$}\label{tab:ablations_alpha}\\[-1pt]
\setlength{\tabcolsep}{2.8pt}
\begin{tabular}{lccccccc}
\hline
$\alpha$ & 0 & 8 & 16 & 32 & 64 & 128 & 256 \\
\hline
$\Delta_m$ (\%) &
-9.90 & +2.77 & +2.77 & +2.75 & +2.82 & +2.79 & +2.81 \\
\hline
\end{tabular}

\end{minipage}

\end{table*}
\vspace{-8pt}

\section{Conclusion}
\label{sec:conclusion}

CRLoMTL combines task-specific LoRA adaptation with shared-subspace fusion, achieving a favorable accuracy--parameter trade-off across three MTL benchmarks without inference-time overhead. Future work will explore larger task settings, other PEFT modules, and shared-subspace dynamics.

\section*{Acknowledgment} \label{ack}
This work is partially supported by NSF grant numbers $2453413$ and $2350180$.

\clearpage
\bibliographystyle{unsrtnat}
\bibliography{main}

\clearpage
\appendix

\begin{center}
  {\Large\bfseries Supplementary Material for\\
  CRLoMTL: Conflict-Resolution through Low-Rank Matrices for Multi-Task Learning}
\end{center}

\section{Additional SVD Resolver Analysis}
\label{supp:svd_analysis}

\begin{figure}[h]
    \centering
    \includegraphics[width=0.75\linewidth]{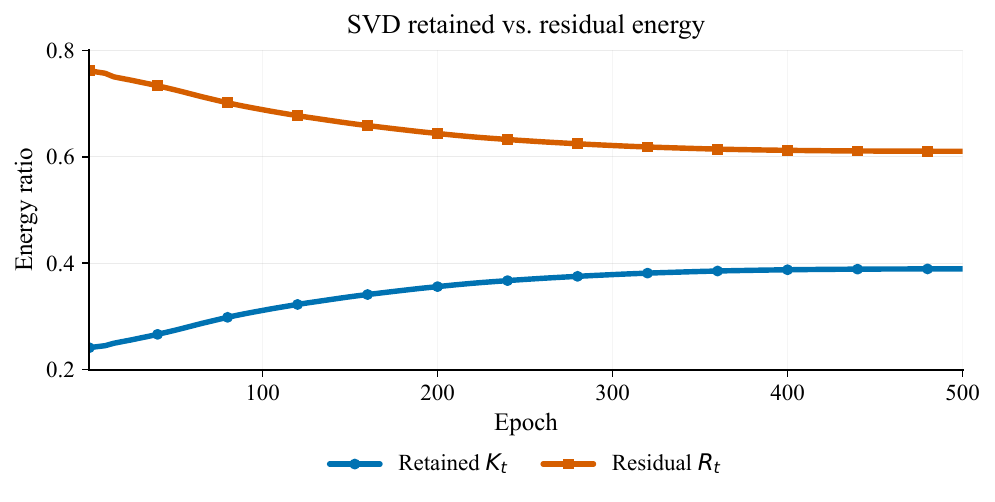}
    \caption{Retained and residual energy over training.}
    \label{fig:app_svd_energy}
\end{figure}

\paragraph{Retained and residual energy:}
We additionally analyze the relative Frobenius energy of the retained component
$\mathbf{K}_{\ell,t}$ and the residual $\mathbf{R}_{\ell,t}$, as shown in
Figure~\ref{fig:app_svd_energy}. The retained component does not necessarily
contain the majority of the update energy; instead, it captures the portion of
the task-specific update represented by the shared SVD basis. In contrast, the
residual may contain a larger fraction of the energy, but it is substantially
less compatible across tasks, as indicated by the NPR analysis in the main paper.
This suggests that the resolver extracts a compact shared subspace while leaving
larger task-specific variation in the residual.

\section{Additional Feature-Space Analysis}
\label{supp:feature_space}

Beyond aggregate metrics, we compare the feature spaces induced by \textsc{CRLoMTL}
and the vanilla MTL baseline. Using t-SNE projections and silhouette scores, we
visualize how each method separates task-specific embeddings while preserving
intra-task cohesion. This qualitative analysis highlights differences in feature
compactness and cross-task interference that are not always apparent from scalar
metrics alone.

\begin{figure}[t]
    \centering
    \begin{minipage}[b]{0.48\linewidth}
        \centering
        \includegraphics[width=\linewidth]{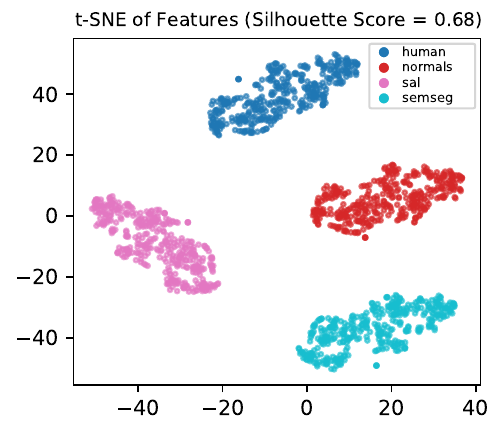}\\
        \small (a) t-SNE of MTL.
    \end{minipage}
    \hfill
    \begin{minipage}[b]{0.48\linewidth}
        \centering
        \includegraphics[width=\linewidth]{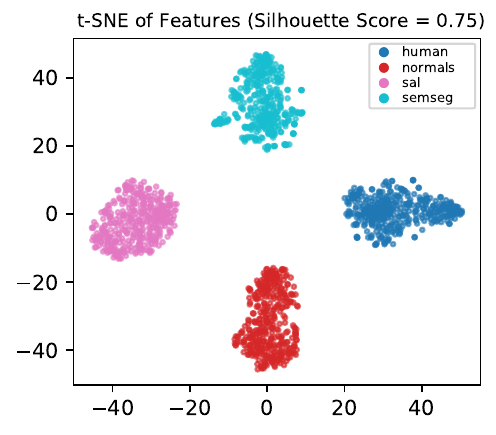}\\
        \small (b) t-SNE of CRLoMTL.
    \end{minipage}
    \caption{t-SNE visualization of task-specific features learned by the vanilla
    MTL baseline and CRLoMTL after 175 training epochs on PASCAL-Context.}
    \label{fig:supp_pascal_tsne}
\end{figure}

Figure~\ref{fig:supp_pascal_tsne} shows t-SNE visualizations of task-specific
features for MTL and CRLoMTL after 175 training epochs on PASCAL-Context. Both
methods exhibit separable features across tasks; however, CRLoMTL produces more
compact clusters per task, indicating stronger intra-task feature consistency.
This observation is further supported by a higher silhouette score for CRLoMTL,
reflecting improved cluster compactness and separation. In contrast, MTL shows
elongated clusters, which may reflect feature noise or redundancy. The silhouette
coefficient measures how well a data point is assigned to its cluster relative to
other clusters~\cite{ROUSSEEUW198753}.

\section{Additional Ablation and Stability Analysis}
\label{supp:additional_ablation}

Table~\ref{tab:supp_seed_stability} provides the per-task variation across three
random seeds. The small standard deviations for semantic segmentation, human
parts, saliency, and surface normals show that the performance gains are not
driven by a single task or a favorable initialization. The overall
$\Delta_m=2.74\pm0.03\%$ further indicates that CRLoMTL maintains a consistent
multi-task improvement across runs.

\begin{table}[h]
\centering
\caption{Seed stability of CRLoMTL on PASCAL-Context over three random seeds.}
\label{tab:supp_seed_stability}
\vspace{4pt}
\setlength{\tabcolsep}{4pt}
\renewcommand{\arraystretch}{0.95}
\begin{tabular}{lccccc}
\hline
Method & SemSeg & Human Parts & Saliency & Normals & $\Delta_m$ \\
       & mIoU $\uparrow$ & mIoU $\uparrow$ & mIoU $\uparrow$
       & RMSE $\downarrow$ & (\%) $\uparrow$ \\
\hline
CRLoMTL
& $67.86\pm0.06$
& $60.43\pm0.02$
& $64.93\pm0.02$
& $16.48\pm0.01$
& $2.74\pm0.03$ \\
\hline
\end{tabular}
\end{table}
\section{CelebA Attribute Grouping}
\label{supp:celeba_groups}

For CelebA, the 40 attributes are grouped manually by semantic similarity, and
the same grouping is used for all runs. The seven groups are:

\begin{itemize}
    \item \textbf{Hair:} Bald, Bangs, Black Hair, Blond Hair, Brown Hair,
    Gray Hair, Receding Hairline, Straight Hair, Wavy Hair.

    \item \textbf{Facial hair:} 5 o'Clock Shadow, Goatee, Mustache, No Beard,
    Sideburns.

    \item \textbf{Expression:} Smiling, Mouth Slightly Open, Narrow Eyes.

    \item \textbf{Makeup:} Arched Eyebrows, Bags Under Eyes, Heavy Makeup,
    Rosy Cheeks, Wearing Lipstick.

    \item \textbf{Accessories:} Eyeglasses, Wearing Earrings, Wearing Hat,
    Wearing Necklace, Wearing Necktie.

    \item \textbf{Face shape:} Attractive, Big Lips, Big Nose, Chubby,
    Double Chin, High Cheekbones, Oval Face, Pointy Nose, Pale Skin.

    \item \textbf{Identity/style:} Blurry, Bushy Eyebrows, Male, Young.
\end{itemize}

\end{document}